\documentclass[preprint,12pt]{elsarticle}

\usepackage{url}
\usepackage{amssymb,epsfig}
\usepackage{graphics,graphicx}
\usepackage{color}
\usepackage{colortbl}
\usepackage{tabularx}

\definecolor{int01}{RGB}{250,250,250}
\definecolor{int02}{RGB}{245,245,245}
\definecolor{int03}{RGB}{240,240,240}
\definecolor{int04}{RGB}{235,235,235}
\definecolor{int05}{RGB}{230,230,230}
\definecolor{int06}{RGB}{225,225,225}
\definecolor{int07}{RGB}{220,220,220}
\definecolor{int08}{RGB}{215,215,215}
\definecolor{int09}{RGB}{210,210,210}

\definecolor{int86}{RGB}{70,70,70}
\definecolor{int87}{RGB}{65,65,65}
\definecolor{int88}{RGB}{60,60,60}
\definecolor{int89}{RGB}{55,55,55}
\definecolor{int90}{RGB}{50,50,50}
\definecolor{int91}{RGB}{45,45,45}
\definecolor{int92}{RGB}{40,40,40}
\definecolor{int93}{RGB}{35,35,35}
\definecolor{int94}{RGB}{30,30,30}
\definecolor{int95}{RGB}{25,25,25}
\definecolor{int96}{RGB}{20,20,20}
\definecolor{int97}{RGB}{15,15,15}
\definecolor{int98}{RGB}{10,10,10}
\definecolor{int99}{RGB}{5,5,5}

\begin{document}

\begin{frontmatter}

\title{Fuzzy Statistical Matrices for Cell Classification}

\author[OHSU]{Guillaume Thibault}
\ead{thibaulg [at] ohsu dot edu}
\author[GOOGLE]{Izhak Shafran}

\address[OHSU]{OHSU Center for Spatial System Biomedicine, BME}
\address[GOOGLE]{Google Inc.}

\begin{abstract}
In this paper, we generalize image (texture) statistical descriptors and propose algorithms that improve their efficacy. Recently, a new method showed how the popular Co-Occurrence Matrix (COM) can be modified into a fuzzy version (FCOM) which is more effective and robust to noise. Here, we introduce new fuzzy versions of two additional higher order statistical matrices: the Run Length Matrix (RLM) and the Size Zone Matrix (SZM). We define the fuzzy zones and propose an efficient algorithm to compute the descriptors. We demonstrate the advantage of the proposed improvements over several state-of-the-art methods on three tasks from quantitative cell biology: analyzing and classifying Human Epithelial type 2 (HEp-2) cells using Indirect Immunofluorescence protocol (IFF).
\end{abstract}


\end{frontmatter}

\section{Introduction}
\label{Sec_Intro}

Human Epithelial Type 2 Cells processed by Indirect Immunofluorescence protocol is the standard method of identifying antinuclear autoantibodies (ANA), and consequently detecting autoimmune diseases such as systemic lupus erythematosus (SLE), rheumatoid arthritis, multiple sclerosis and diabetes~\cite{ERHW10,MS10,WHFCM10}. However, current methods require at least one expert to visually analyze the distributions of antibodies across multiple images. Usually this analysis is performed through a microscope and is comprised of three steps~\cite{CBIFV14}: i) detection of at least one mitotic cell, ii) evaluation of the fluorescence signal intensity (negative in the absence of fluorescence, else intermediate or positive), iii) determining the cells classification according to the auto-antibody type distribution. These multi-steps manual analyses are tedious, time consuming, subjective and have high inter-/intra-observer variability~\cite{HBKR09} (up to $24\%$, as reported in~\cite{BTTPM98,FPSV13}). Moreover, the increasing number of patients and the limited number of experts make this impractical to scale to a large number of clinics. Therefore, a stable and effective automatic Computer-Aided Diagnosis (CAD) system is needed.\\
Hopefully, cell classification is now a well-established task~\cite{CJL06,PPMu02}, as the advent of high-throughput imaging techniques has introduced the need for a robust system to automatically analyze thousands of cell images~\cite{NW10}. Typically, most classification systems consist of two cascaded modules -- one module that extracts useful features from a cell or a group of cells, followed by a second module that classifies the cells or the group using the extracted features. Unfortunately, the range of images qualities as well as the classes to predict (see Fig.~\ref{Figure_Cells0},~\ref{Figure_Cells1} and~\ref{Figure_Cells2}) makes cell classification a particularly complicated task.\\

In this paper, we address these imaging issues by introducing new texture features extraction methods. These methods are robust to quality variations (particularly noise), and able to efficiently describe a wide variety of classes. This was accomplished by introducing fuzzy logic before the filling of statistical matrices. In order to demonstrate that our work can be used for different cytology purposes, we use three datasets composed of IFF images, which contain different image qualities as well as classes to predict.\\
Before delving into the paper, we first describe the three representative tasks from quantitative image-based cell biology. Next we outline a typical cell classification system (section \ref{Section_Classification}), and present a review of the different statistical matrices (section \ref{Section_StatisticalMatrices}). Then we present our work: a fuzzy generalization of existing statistical matrices (section \ref{SubSection_PreviousFuzzy}), as well as the fuzzy zone definition and computation (section \ref{SubSection_FuzzyZones}). Finally, the proposed matrices are evaluated on three tasks for classifying cells and their structures (section~\ref{Section_Results}).\\

\newpage

\section{Datasets}
\label{Sec_data}

\begin{figure}[!ht]
\centering
\begin{tabular}{ccc}
	\includegraphics[scale=0.75]{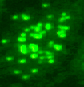}		&	\includegraphics[scale=0.825]{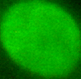}	&	\includegraphics[scale=1.125]{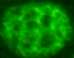}\\
	\small{Centromere (357)}								&	\small{Homogeneous (330)}							&	\small{Coarse Speckles (210)}\\
	\includegraphics[scale=0.75]{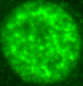}		& 	\includegraphics[scale=0.775]{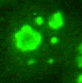}	&	\includegraphics[scale=0.35,angle=90]{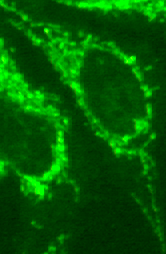}\\
	\small{Fine Speckles (208)}							&	\small{Nucleolar (241)}								&	\small{Cytoplasmic (110)}\\
\end{tabular}
\caption{Example of cell images from the ICPR 2012 Cell Classification Contest. The number of cells per class is indicated within the parentheses.}
\label{Figure_Cells0}
\end{figure}

\noindent\textbf{ICPR 2012 HEp-2 Cells Classification Contest -} This widely used dataset~\cite{CBIFV14,FPSV13,NF14,FHWL14,SMS14,SLWY14,YWALH14,NPB14} is composed of $1456$ cells manually segmented from $28$ IFF images, and annotated by experts. Each image contains many cells (min $13$, max $119$, with average dimensions about $86 \times 87$ pixels) of a unique type, which can be one of the six imbalanced classes (see Fig.~\ref{Figure_Cells0}): Centromere (CE), uniform discrete speckles located throughout the entire nucleus; Homogeneous (HO), diffuse staining in the entire nucleus; Coarse Speckles (CS), densely distributed, variously sized speckles, generally associated with larger speckles; Fine Speckles (FS), fine speckled staining in an uniform distribution, sometimes very dense and almost homogeneous; Nucleolar (NU), less than six large coarse speckled staining within the nucleus; Cytoplasmic (CY), fine dense granular to homogeneous staining or cloudy pattern, covering part or the whole cytoplasm.
\\

\begin{figure}[!ht]
\centering
\begin{tabular}[t]{ccc}
	\includegraphics{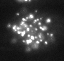} & \includegraphics[scale=0.925]{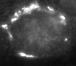} & \includegraphics[scale=0.975]{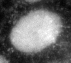}\\
	\small{Centromere}  & \small{Golgi}  & \small{Homogeneous}\\
	% & & \\
	\includegraphics[angle=90]{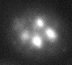} & \includegraphics{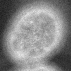} & \includegraphics{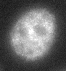}\\
	\small{Nucleolar} & \small{NuMem} & \small{Speckled}\\
\end{tabular}
\caption{Example of cell images from the ICIP 2013 Cell Classification contest.}
\label{Figure_Cells1}
\end{figure}

\noindent\textbf{ICIP 2013 Cell Classification Contest -} The dataset is comprised of more than $13500$ cells categorized into 6 classes (see Fig.~\ref{Figure_Cells1}): Centromeres (CE), NuMem (NM), Speckled (SP), Golgi (GO), Homogeneous (HO) and Nucleolar (NU). Apart from accuracy, the task  evaluates robustness with images in two conditions: "positive" condition with normal illumination, and "intermediate" condition with high levels of noise, under exposed or low contrasted images (with a narrowed histogram concentrated on the left). These variations in noise and contrast make this dataset a good candidate to evaluate the methods being studied in this paper.

\begin{figure}[!ht]
\center
\begin{tabular}{cccc}
	\includegraphics[scale=0.175]{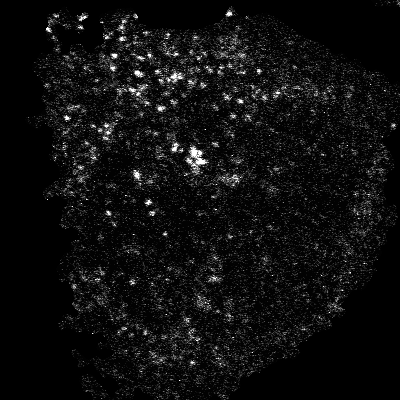} & \includegraphics[scale=0.175]{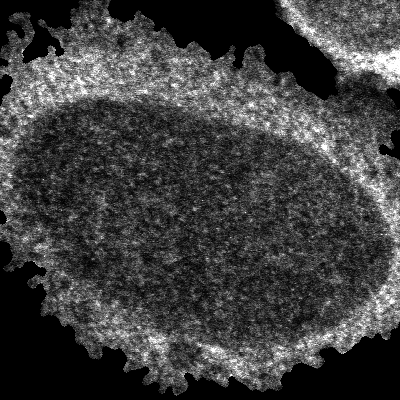} & \includegraphics[scale=0.175]{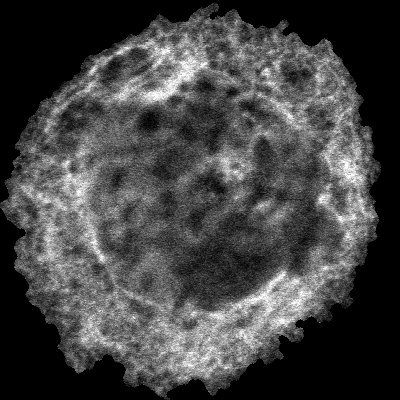} & \includegraphics[scale=0.175]{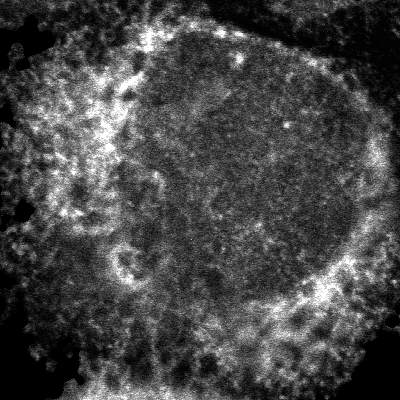}\\
	\small{Centrosome ($32$)} & \small{Cytoplasm ($144$)} & \small{Cytoskeleton ($38$)} & \small{ER ($42$)}\\
	& & & \\
	 \includegraphics[scale=0.175]{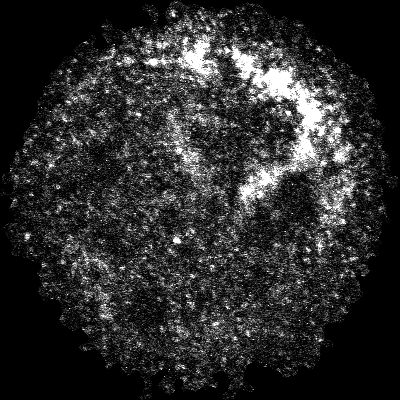} & \includegraphics[scale=0.175]{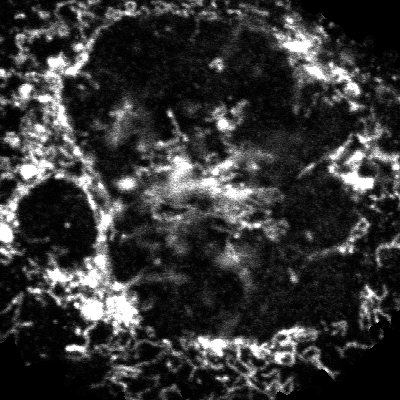} & \includegraphics[scale=0.175]{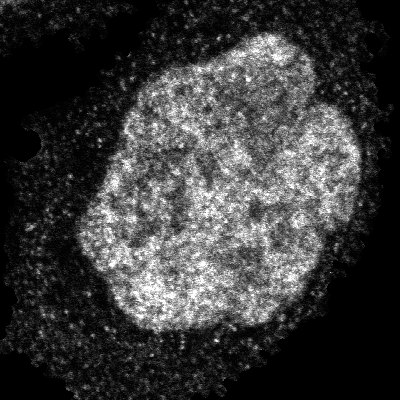} & \includegraphics[scale=0.175]{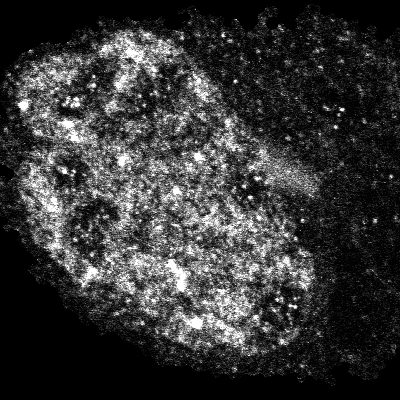}\\
	 \small{Golgi ($64$)} & \small{Mitochondria ($180$)} & \small{Nuclei ($96$)} & \small{Nuclei w/o ($470$)}\\
	 & & & \\
	\includegraphics[scale=0.175]{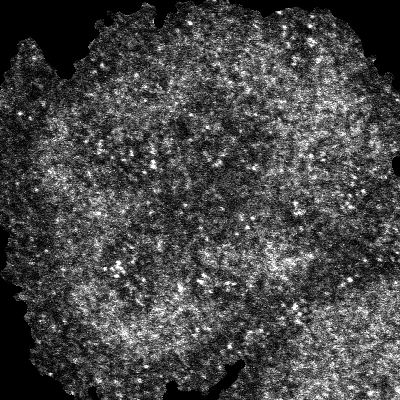} & \includegraphics[scale=0.175]{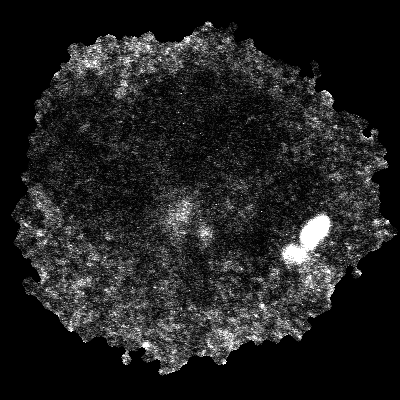} & \includegraphics[scale=0.175]{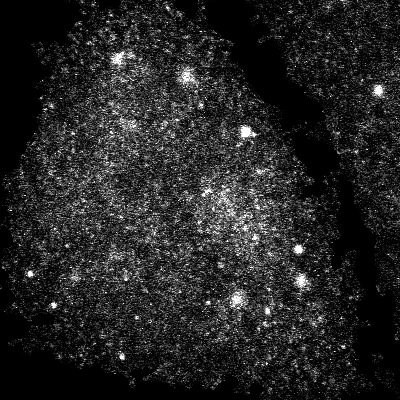} & \\
	\small{Nucleoli ($244$)} & \small{Plasma ($34$)} & \small{Vesicles ($140$)} & \\
\end{tabular}
\caption{Example of cell images from Human Protein Atlas (HPA).}
\label{Figure_Cells2}
\end{figure}

\noindent\textbf{Cell Protein Classification in HPA IF Images -} This task is comprised of IFF images from the Human Proteins Atlas (HPA) project~\cite{UOFLJ10,LNULM12} that show sub-cellular locations for thousands of proteins. Images were annotated by visual inspection and classified into $11$ classes by experts (cf. Fig.~\ref{Figure_Cells2}). Of the $1484$ images, a subset of images containing a single class per image, all with good staining qualities were culled to create our evaluation test set. The images were segmented using mathematical morphology and automatic thresholding. This dataset contains high quality images, but the class distributions are skewed.

\section{Classification}
\label{Section_Classification}

The task in a typical cell-classification problem is to assign a class label to each input image. An image may consist of one or more cells, in which case the task becomes more complicated, involving a segmentation stage that occurs either separately or simultaneously. In a supervised classification scenario, a set of example images with reference labels is available to train the classifier or learn models. Prior to the classification phase, as mentioned before, useful features need to be extracted from the input image, often in the form of a vector. The greater the feature vector, the higher the capacity of the model, which often results in better classification accuracy. However, this may allow the model to memorize the training data, and as a result may generalize poorly to test inputs that are not well represented in the training data. This is more acute when the training data is limited in size or diversity. We alleviate this problem by adopting \textit{K-Fold Cross Validation}~\cite{Dietterich98,Kohavi95,Stone74} in our experimental evaluations.\\
In this paper we consider two popular and effective classification methods from machine learning:
\begin{itemize}
	\item Random Forests (RF)~\cite{Breiman01} are non-linear classifiers and are based on \textit{Classification And Regression Trees} (CART)~\cite{Breiman84}, where the decisions at each node are randomized in a manner that improve generalization~\cite{Breiman96,AmitGeman97}.
	\item Neural Networks~\cite{ERH02} (NN) are non-linear classifiers comprising of a collection of nodes that attempt to imitate the distributed computing of the neurons in brain. The parameters of the nodes are learned automatically from the data using back propagation of errors incurred in the cost function (e.g., average squared error, cross entropy).
\end{itemize}

In many cell-classification problems, there may be multiple cells or segments in the images, where all segments may not have the same label. Such problems require segmenting the image and then classifying each segment separately. Classification of images or segments into one of many ($N$) classes is typically solved using $N$ binary classifiers, where each classifier differentiates one unique class from the rest. In most natural tasks, the distributions of classes are skewed, and are rarely uniform. This poses additional problems for training a multi-class classifier. A number of techniques are available to mitigate this, including over/under sampling~\cite{LWZ06} (random or directed addition/suppression of instances in the minority/majority class until the sets are balanced), methods based on asymmetric entropy measure~\cite{MZR06} and auto-associator neural networks~\cite{Japkowicz00}. We adopt a re-weighting scheme that increases the cost associated with errors from infrequent classes and evaluate our algorithms against random chance.

\section{Previous Works on Statistical Matrices}
\label{Section_StatisticalMatrices}

Let $f:\left\{\begin{array}{c}E \rightarrow {\cal T}\\\mathbf{x} \mapsto f(\mathbf{x}) \end{array}\right.$ be a gray-levels image with dimensions $w \times h$, where $E\subset \mathbf{Z}^{2}$ is the pixels support space and the image intensities are discrete values which range in a closed set ${\cal T}=\{t_1, t_2, ..., t_N\}$, $\Delta t = t_{i+1}-t_{i}$, e.g., for an $8$ bits image $t_1=1$, $N=256$ and $\Delta t=1$. Assume that the image $f$ is segmented into its $J$ flat zones $R_j[f]$ (i.e., connected regions of constant value): $E = \cup_{j=1}^{J} R_{j}[f]$, $\cap_{j=1}^{J} R_{j}[f] =\emptyset$. Each region size (surface area) is $s(j) = |R_{j}[f]|$ ($|.|$ is the cardinal). Hence, we consider that each zone $R_j[f]$ has an associated constant gray-level intensity.\\

Statistical matrices have been extensively used in texture characterization, the best known of which is the gray level Co-Occurrence Matrix (COM), which leads to the definition of Haralick's features~\cite{Haralick73}. The COM represents the texture by second order statistics: co-occurring values distribution at a given offset. For an offset $\Delta=(\Delta_x, \Delta_y)$, the COM is defined as:
$$
COM_{f, \Delta}(i,j) = \sum_{x=1}^{w} \sum_{y=1}^{h} \left\{
\begin{array}{l}
	1, \; if \; f(x,y)=i \; and \; f(x+\Delta_x, y+\Delta_y) = j \\
	0, \; otherwise
\end{array}
\right.
$$
By design, the COM is dependent on the offset and therefore is not rotation invariant. When using 8-connexity, this is addressed by computing the COM in four directions with the offsets $\theta_{0^\circ}=(0,1)$, $\theta_{45^\circ}=(1,1)$, $\theta_{90^\circ}=(1,0)$, $\theta_{135^\circ}=(-1,1)$, and then the average matrix over all offsets can be used~\cite{AlKadi10,JEC00,WZBCPW08}. The amount of information extracted depends on the number of offset directions and their norm. Typically, a large number of offsets are needed to extract all the useful information, which is the main drawback of this approach.

Second order statistics can also be extracted with:
\begin{itemize}
	\item  The gray level Difference Histogram (DH)~\cite{DCA79,Unser86,SCD95}, an absolute differences histogram, defined as:
	$$ DH_{f,\Delta} (i) = \sum_{x=1}^{w} \sum_{y=1}^{h} \left\{ \begin{array}{l} 1, \; if \; |f(x,y)-f(x+\Delta_x, y+\Delta_y)|=i\\0, \; else \end{array} \right. $$ %\begin{array}{l} 1, \; if \; |f(x,y)-\\ \hspace{0.9cm}f(x+\Delta_x, y+\Delta_y)|=i\\0, \; else \end{array} \right. $$
	\item The gray level Sum Histogram (SH)~\cite{Unser86,SCD95}:%, defined as:
	$$ DH_{f,\Delta} (i) = \sum_{x=1}^{w} \sum_{y=1}^{h} \left\{ \begin{array}{l} 1, \; if \; f(x,y) + f(x+\Delta_x, y+\Delta_y)=i\\0, \; else \end{array} \right. $$
\end{itemize}
These methods extract less information than COM. However, Weska et al. found that they provide performances similar to Haralick's features in some applications~\cite{WDR76}.

Another classical technique is the gray level Run Length Matrix (RLM)~\cite{Galloway75}, which has been extensively used for texture classification~\cite{Chu90,Dasarathy91}. The RLM extracts higher order statistical features: the matrix element $RLM_{f,\theta}(g,l)$ counts the number of runs (i.e., collinear pixels with the same intensity in the direction $\theta$) with the gray level $g$ and length $l$ (see Fig.~\ref{Figure_RLM}). This method is particularly effective for periodic textures and completes the information provided by the COM. Extracted features from the RLM are moments of order from $-2$ to $2$.
\begin{figure}[!ht]
	\centering
	\small
	\begin{tabular}{cc}
		{\renewcommand{\arraystretch}{1.5}
		\begin{tabular}{|c|c|c|c|}
			\hline
			\cellcolor{black}{\textcolor{white}{$1$}} &	\cellcolor[RGB]{85,85,85}{\textcolor{white}{$2$}}	&	\cellcolor[RGB]{170,170,170}{$3$}			&	$4$\\
			\hline
			\cellcolor{black}{\textcolor{white}{$1$}} &	\cellcolor[RGB]{170,170,170}{$3$}				&	$4$									&	$4$\\
			\hline
			\cellcolor[RGB]{170,170,170}{$3$}	&	\cellcolor[RGB]{85,85,85}{\textcolor{white}{$2$}}	&	\cellcolor[RGB]{85,85,85}{\textcolor{white}{$2$}}	&	\cellcolor[RGB]{85,85,85}{\textcolor{white}{$2$}}\\
			\hline
			$4$							&	\cellcolor{black}{\textcolor{white}{$1$}}			&	$4$									&	\cellcolor{black}{\textcolor{white}{$1$}}\\
			\hline
		\end{tabular}}
		&
		\begin{tabular}{|c|c|}
			\hline
			\textit{Level} & \textit{Run length}, $l$\\
			$g$ & \begin{tabular}{c|c|c} 1 & 2 & 3 \end{tabular}\\
			\hline
			1 & \begin{tabular}{ccc} 4 & 0 & 0 \end{tabular} \\
			2 & \begin{tabular}{ccc} 1 & 0 & 1 \end{tabular} \\
			3 & \begin{tabular}{ccc} 3 & 0 & 0 \end{tabular} \\
			4 & \begin{tabular}{ccc} 3 & 1 & 0 \end{tabular} \\
			\hline
		\end{tabular}\\
		(a) & (b)
	\end{tabular}
\caption{RLM filling example for a $4$ gray levels image texture of size $4 \times 4$, with $\theta=0^\circ$.}
\label{Figure_RLM}
\end{figure}
\normalsize

Recently Thibault et al.~\cite{Thibault09,Thibault13,ThibaultIEEE14} introduced the gray level Size Zone Matrix (SZM) original notion, as an alternative to the joint RLM distribution. The SZM is based on each flat zone size/intensity co-occurrences, and therefore provides a statistical representation by the bivariate conditional probability density function estimation of the image distribution values. In this method, the matrix value $SZM_{f}(s,g)$ counts the number of zones with a size $s$ and a gray level $g$ in $f$ (see Fig.~\ref{Figure_SZM}). The resulting matrix has a fixed number of rows equal to $t_N$ (the gray level number, determining the matrix's height), and a dynamic number of columns (the matrix's width), determined by the largest zone size as well as the size quantization. The image gray levels number (resp. sizes) can be reduced by a function in order to improve results efficiency and stability. In this matrix, the more homogeneous the texture (large flat zones with closed gray levels), the wider and flatter the matrix. From this statistical matrix representation, we can calculate all the second-order moments as compact texture features~\cite{Chu90}, plus two features which are specific weighted variances~\cite{Thibault09}.

\begin{figure}[!ht]
	\centering
	\small
	\begin{tabular}{cc}
		{\renewcommand{\arraystretch}{1.5}
		\begin{tabular}{|c|c|c|c|}
			\hline
			\cellcolor{black}{\textcolor{white}{$1$}} &	\cellcolor[RGB]{85,85,85}{\textcolor{white}{$2$}}	&	\cellcolor[RGB]{170,170,170}{$3$}			&	$4$\\
			\hline
			\cellcolor{black}{\textcolor{white}{$1$}} &	\cellcolor[RGB]{170,170,170}{$3$}				&	$4$									&	$4$\\
			\hline
			\cellcolor[RGB]{170,170,170}{$3$}	&	\cellcolor[RGB]{85,85,85}{\textcolor{white}{$2$}}	&	\cellcolor[RGB]{85,85,85}{\textcolor{white}{$2$}}	&	\cellcolor[RGB]{85,85,85}{\textcolor{white}{$2$}}\\
			\hline
			$4$							&	\cellcolor{black}{\textcolor{white}{$1$}}			&	$4$									&	\cellcolor{black}{\textcolor{white}{$1$}}\\
			\hline
		\end{tabular}}
		&
		\begin{tabular}{|c|c|}
			\hline
			\textit{Level} & \textit{Size zone}, $s$\\
			$g$ & \begin{tabular}{c|c|c} 1 & 2 & 3 \end{tabular}\\
			\hline
			1 & \begin{tabular}{ccc} 2 & 1 & 0 \end{tabular} \\
			2 & \begin{tabular}{ccc} 1 & 0 & 1 \end{tabular} \\
			3 & \begin{tabular}{ccc} 0 & 0 & 1 \end{tabular} \\
			4 & \begin{tabular}{ccc} 2 & 0 & 1 \end{tabular} \\
			\hline
		\end{tabular}\\
		(a) & (b)
	\end{tabular}
\caption{SZM filling example for a $4$ gray levels image texture of size $4 \times 4$ and using $8$-connexity.}
\label{Figure_SZM}
\end{figure}
\normalsize

Unlike COM and RLM, which dependent on the offset $\Delta$ and the orientation $\theta$ respectively, the SZM is invariant with respect to rotation and translation. However, it requires a flat zone labeling that is time consuming. The connectivity type used for labeling modifies the matrix but does not impact the classification performances~\cite{ThibaultIEEE14}. RLM and COM are appropriate for periodic textures whereas the SZM is typically adapted to describe heterogeneous non-periodic textures. In addition, due to the intrinsic segmentation, texture description in SZM is more regional than the point-wise-based COM representation.\\

There are several variants of the SZM~\cite{ThibaultICIP11,ThibaultIEEE14}. One of them is the Multiple gray level SZM (MSZM), which is computed from $N$ SZM for $N$ different gray levels quantizations $\{N_1, ... N_N\}$. The resulting matrices are combined by a weighted average: $ MSZM_{f}(s,g) = \sum_{k=1}^{N}w_{k} SZM_{f}^{N_k}(s,g) $. Two other SZM variants are specially designed to characterize specific biological structures: the microtubule network organization (the gray level Orientation and geodesic Length Zone Matrix, OLZM) and the DNA during mitosis (the gray level Distance-to-border Zone Matrix, DZM). They are effective in certain applications, but are not used in this paper.\\

%Indeed, a pixel gray level variation of $\Delta v$, involves: a different increment located at a distance $\Delta v$ in the COM; two potentially shorter runs (because they are cut) in the RLM; a singleton (flat zone of size $1$) and a zone reduced of $1$ for the SZM.
\textbf{Remark -} By design all these matrices are sensitive to noise (every acquisition devices introduce noise, generally gaussian, during imaging). In order to improve their noise robustness, the texture gray levels number is reduced to $N$ possible values before matrix filling using one of the following method:
\begin{itemize}
	\item A function. First a histogram spreading is first performed, and then a function is applied. Most of the time the function is linear (so a simple division is performed), but $logarithm$ or other functions can be used.
	\item A cumulated histogram in order to separate the pixels distribution into $N$ bins containing approximately the same pixels number.
	\item A clustering algorithm with $N$ clusters.
	\item A dynamic programming, based on Bayesian blocks applied to the image histogram~\cite{QSJ07}.
	\item A combination of the Growing Neural Gas (GNG) and the Kohonen Self-Organized Map (SOFM)~\cite{AP05}.
\end{itemize}
The classification performances can be greatly impacted by the algorithm used, so it is generally recommended to test some or all of them, with different gray level quantizations.

\section{Fuzzy Boundaries}
\label{Section_fuzzification}

\subsection{Previous Work: Fuzzy Co-occurrence Matrix}
\label{SubSection_PreviousFuzzy}

\begin{figure}[!ht]
\centering
\includegraphics[scale=0.4]{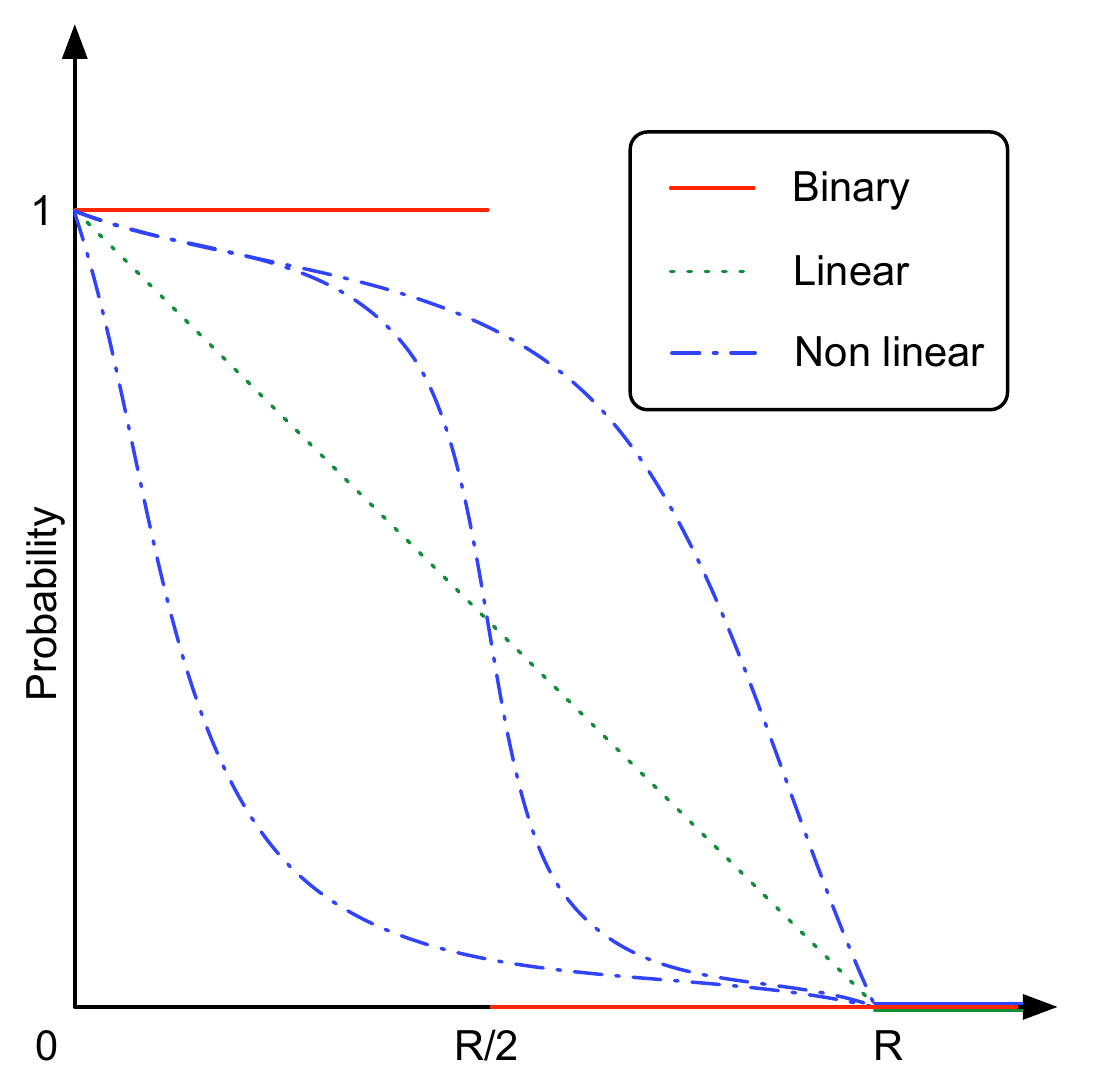}
\caption{Examples of membership functions: binary (red), linear (green) and non linear (blue).}
\label{Figure_FuzzyFunctions}
\end{figure}

In~\cite{SP06} authors use fuzzy logic principles to introduce a COM fuzzy version. In the original version, each pixels pair $(i,j)$ increases $COM_{f,\Delta}(i,j)$ by $1$. The fuzzy version uses a membership function $\beta$, which is a real monotonically decreasing probability function, with a fuzzy parameter $R$ being the neighborhood radius (see Fig.~\ref{Figure_FuzzyFunctions}). The membership function is used to increase the fuzzy co-occurrence matrix $FCOM_{f,\Delta}(i,j)$ and its neighborhood. Therefore the FCOM gives the gray values occurrence frequency around a value $s$ located at an offset $\Delta$ around another gray level value $t$. According to the authors, this decreases the COM noise sensitivity.\\
This principle can be immediately applied to the SZM (resp. RLM), which we refer as FSZM (resp. FRLM):  $FSZM_{f,\beta}(s,g)$ represents the sum of all the probabilities for a zone of size $s$ and gray level $g$ to exist in $f$.

\subsection{New Fuzzy Versions Using Fuzzy Zones}
\label{SubSection_FuzzyZones}

\begin{figure}[!ht]
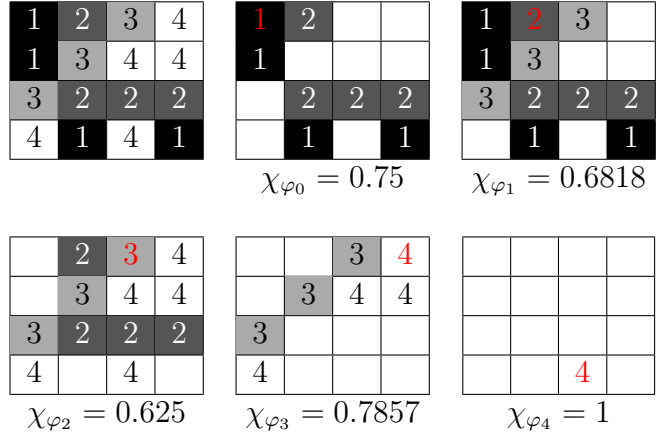

\centering
\begin{tabular}{ccc}
\begin{tabular}{|c|c|c|c|}
	\hline
	\cellcolor{black}{\textcolor{white}{$1$}} 	&	\cellcolor[RGB]{85,85,85}{\textcolor{white}{$2$}}	&	\cellcolor[RGB]{170,170,170}{$3$}				&	$4$\\
	\hline
	\cellcolor{black}{\textcolor{white}{$1$}} 	&	\cellcolor[RGB]{170,170,170}{$3$}				&	$4$										&	$4$\\
	\hline
	\cellcolor[RGB]{170,170,170}{$3$}		&	\cellcolor[RGB]{85,85,85}{\textcolor{white}{$2$}}	&	\cellcolor[RGB]{85,85,85}{\textcolor{white}{$2$}}	&	\cellcolor[RGB]{85,85,85}{\textcolor{white}{$2$}}\\
	\hline
	$4$								&	\cellcolor{black}{\textcolor{white}{$1$}}			&	$4$										&	\cellcolor{black}{\textcolor{white}{$1$}}\\
	\hline
\end{tabular}	&
\begin{tabular}{|c|c|c|c|}
	\hline
	\cellcolor{black}{\textcolor{red}{$1$}} &	\cellcolor[RGB]{85,85,85}{\textcolor{white}{$2$}}	&				&	\\
	\hline
	\cellcolor{black}{\textcolor{white}{$1$}} &					&										&	\\
	\hline
		&	\cellcolor[RGB]{85,85,85}{\textcolor{white}{$2$}}	&	\cellcolor[RGB]{85,85,85}{\textcolor{white}{$2$}}	&	\cellcolor[RGB]{85,85,85}{\textcolor{white}{$2$}}\\
	\hline
								&	\cellcolor{black}{\textcolor{white}{$1$}}			&										&	\cellcolor{black}{\textcolor{white}{$1$}}\\
	\hline
\end{tabular}	&
\begin{tabular}{|c|c|c|c|}
	\hline
	\cellcolor{black}{\textcolor{white}{$1$}} &	\cellcolor[RGB]{85,85,85}{\textcolor{red}{$2$}}	&	\cellcolor[RGB]{170,170,170}{$3$}			&	\\
	\hline
	\cellcolor{black}{\textcolor{white}{$1$}} &	\cellcolor[RGB]{170,170,170}{$3$}				&										&	\\
	\hline
	\cellcolor[RGB]{170,170,170}{$3$}	&	\cellcolor[RGB]{85,85,85}{\textcolor{white}{$2$}}	&	\cellcolor[RGB]{85,85,85}{\textcolor{white}{$2$}}	&	\cellcolor[RGB]{85,85,85}{\textcolor{white}{$2$}}\\
	\hline
								&	\cellcolor{black}{\textcolor{white}{$1$}}			&										&	\cellcolor{black}{\textcolor{white}{$1$}}\\
	\hline
\end{tabular}\\
 & $\chi_{\varphi_0}=0.75$ & $\chi_{\varphi_1}=0.6818$\\
 & & \\
\begin{tabular}{|c|c|c|c|}
	\hline
	 &	\cellcolor[RGB]{85,85,85}{\textcolor{white}{$2$}}	&	\cellcolor[RGB]{170,170,170}{\textcolor{red}{$3$}}			&	$4$\\
	\hline
	 &	\cellcolor[RGB]{170,170,170}{$3$}				&	$4$									&	$4$\\
	\hline
	\cellcolor[RGB]{170,170,170}{$3$}	&	\cellcolor[RGB]{85,85,85}{\textcolor{white}{$2$}}	&	\cellcolor[RGB]{85,85,85}{\textcolor{white}{$2$}}	&	\cellcolor[RGB]{85,85,85}{\textcolor{white}{$2$}}\\
	\hline
	$4$							&				&	$4$									&	\\
	\hline
\end{tabular}	&
\begin{tabular}{|c|c|c|c|}
	\hline
	 &		&	\cellcolor[RGB]{170,170,170}{$3$}			&	\textcolor{red}{$4$}\\
	\hline
	 &	\cellcolor[RGB]{170,170,170}{$3$}				&	$4$									&	$4$\\
	\hline
	\cellcolor[RGB]{170,170,170}{$3$}	&		&		&	\\
	\hline
	$4$						&				&			&	\\
	\hline
\end{tabular}	&
\begin{tabular}{|c|c|c|c|}
	\hline
	\hspace{0.2cm}	& \hspace{0.2cm}	&	& \hspace{0.2cm}	\\
	\hline
	&	&	&	\\
	\hline
	&	&	&	\\
	\hline
	&	&	\textcolor{red}{$4$}	&	\\
	\hline
\end{tabular}\\
$\chi_{\varphi_2}=0.625$ & $\chi_{\varphi_3}=0.7857$ & $\chi_{\varphi_4}=1$
\end{tabular}
\caption{Examples of all possible fuzzy zones from the top left texture, computed with $8$-connexity and a linear membership function $\beta : x \mapsto \max (-\frac{|x|}{2}+1, 0)$. The starting pixel are colored in red.}%, and green pixels could have generated exactly the same fuzzy zone.}
\label{Figure_FuzzyZones}
\end{figure}

% If $\beta$ is binary with $F=1$ (see figure \ref{Figure_FuzzyFunctions}), we get the flat zone definition back $|f(p_0)-f(p_i)| = 0$.\\
The previous fuzzy method introduces the fuzzy part (fuzzification) during the matrix filling, so it still uses the exact values for the texture under study. But as explained in the section \ref{Section_StatisticalMatrices} \textit{Remark}, even the best acquisition device provides only an approximation of the reality, and as a flat zone has a rigid definition (connected pixels set with exactly the same gray level), it is noise sensitive. We tackle this using the fuzzy logic during the texture analysis, and consider each pixel as a fuzzy pixel with a fuzzy gray level, and then we introduce the \textit{fuzzy zone} notion: for an image $f$, a starting pixel $p_0$ and a membership function $\beta$, the fuzzy zone $\varphi$ is formed by all the connected pixels $\{p_0, ..., p_i\}$ such as $\beta(|f(p_0)-f(p_i)|) > 0$. Consequently, the fuzzy zone $\varphi$ is described with:
\begin{enumerate}
	\item The original pixel $p_0$ and its gray level $f(p_0)$.
	\item The pixels constituent $\varphi = \{p_0, ..., p_i\}$, and the associated probabilities $\chi_{p_i}=\beta(|f(p_0)-f(p_i)|)$.
	\item A probability $\chi_\varphi$ computed from the $\chi_{p_i}$ (average, median, etc.).
\end{enumerate}
So the bigger the difference between $f(p_0)$ and the $f(p_i)$ the lower the probability. For example, for a flat zone, $\forall i, \; f(p_0)=f(p_i) \Rightarrow \varphi_{p_i}=1$ and then $\beta_\varphi=1$.  By definition, a pixel can be part of different fuzzy zones, and consequently two fuzzy zones can have exactly the same pixels, but different starting points and probabilities. Moreover, the higher the fuzzy parameter (the membership function parameter $R$), the greater the size while reducing the fuzzy zones number. Figure~\ref{Figure_FuzzyZones} shows an example.% of all the fuzzy zones and their probabilities from a given texture.

All the fuzzy zones can be characterized and used to fill a SZM (or a RLM): for an image $f$ and a fuzzy zone $\varphi$, the size $s(\varphi)$ is computed and the matrix case $SZM_{f,\beta}(f(p_0),s(\varphi))$ is increased by $\beta_\varphi$. The fuzzy zones computation allows to introduce the fuzziness at the image level instead of the matrix filling level. Such a new fuzzy SZM and RLM are annotated \textit{FuzzySZM} and \textit{FuzzyRLM} respectivily. It is no longer required to reduce the gray levels number, and therefore the matrix's height is equal to the image gray levels number. The algorithm required to find the fuzzy zones has a non linear complexity that depends on the fuzzy parameter $R$, and consequently the FuzzySZM/FuzzyRLM filling is much more time consuming (by at least a factor of $5$) than a classical SZM/RLM.\\
%\textbf{Remark -}
This fuzzy version using fuzzy zones fills a matrix with a fixed height equal to the gray levels number in the image. Therefore, the multiple gray levels principle described at the end of section \ref{Section_StatisticalMatrices} no longer makes sense. However, the FuzzySZM required a fixed fuzzy parameter, so a Multiple Fuzzy SZM can be created: the same matrix is filled using different fuzzy parameters.

\section{Results}
\label{Section_Results}

This section presents the results obtained from the three different datasets introduced in section~\ref{Sec_Intro}. All the classic statistical matrices are used with two gray level reduction algorithms (linear and histogram), six quantizations (dyadic values from $8$ to $256$), and our new fuzzy statistical matrices were tested with a linear membership function and for different fuzzy parameters. For each method, only the best result is reported. The blue numbers indicate that the fuzzy version improves the corresponding basic algorithm (COM, RLM, SZM) performances, and the red number points out the optimal performance for each class.\\
In this section, the two classifiers used are: 1) a neural network of type perceptron, with one hidden layer containing $11$ neurons (best configuration experimentally found), trained with back-propagation, using individual adaptive learning rates and double momentums~\cite{PMV01}; 2) random forests with $5$ times more trees than features. Each classifier is then validated using leave-one-out or k-fold cross validation.\\

%\textbf{Remark -}
In section~\ref{SubSection_PreviousFuzzy}, we presented FRLM and FSZM, extensions of RLM and SZM according to the COM fuzzy principle described in~\cite{SP06}. Unfortunately, among the three datasets used in this paper, FRLM and FSZM never improve RLM and SZM performances. Moreover, FCOM slightly improved the COM performances only once, with a $0.14$ gap for the class nucleolar into the ICIP 2013 contest dataset, using the random forests. Consequently,  FCOM, FRLM and FSZM results are not presented in this section, because of lack of efficacy.

\subsection{ICPR 2012 contest dataset}
\label{SubSection_ResultsICPR2012}

The highly reliable and widely used leave-one-out cross-validation was performed over all $28$ images. As each image contains only one type of cell, two different results levels were reported: at the cell level the results try to predict each cell class, and at the image level the results try to predict the most frequently assigned cell class within that image. A six-classes classifier was built using a neural network (lower results were obtained with random forests, and then were not reported), where the results are displayed in tables~\ref{Table_ResultsICPR2012comp} and~\ref{Table_ResultsICPR2012}. The fuzzy versions results were compared with those obtained from the original versions, and with many methods from the state-of-the-art (table \ref{Table_ResultsICPR2012comp}). These methods used different features (such as local binary patterns, morphological, statistical, Fisher tensors, moments, etc.) and classifiers (mainly support vector machines, but neural networks and random forests as well). Contrary to our fuzzy matrices that provide around $15$ features, all these methods use a huge number of features and often require a features selection. But the results show that few highly relevant features can outperform other methods using a large number of features, which demonstrates the efficacy of our fuzzy versions.

\begin{table}[!ht]
\resizebox{\columnwidth}{!}{
\begin{tabular}{cc}
\begin{tabular}{|c|cccccc|}
	\hline
	Methods		&	CE					&	HO					&	CS					&	FS					&	NU					&	CY	\\
	\hline
	\hline
	FuzzySZM	& 	\textcolor{blue}{$93.2$}	&	$92.1$				&	\textcolor{blue}{$93.3$}	&	$92.3$				&	\textcolor{red}{$97.9$}	&	\textcolor{red}{$100$}\\
	SZM			& 	$91.5$				&	$92.8$				&	$92.4$				&	\textcolor{red}{$95.8$}	&	$92.4$				&	$99$\\
	\hline
	\hline
	FuzzyRLM	& 	\textcolor{red}{$94.4$}	&	\textcolor{red}{$94.2$}	&	\textcolor{red}{$94.7$}	&	$92.7$				&	\textcolor{blue}{$97.5$}	&	\textcolor{red}{$100$}\\
	RLM			& 	$84.2$				&	$91.8$				&	$91.6$				&	$94.6$				&	$94.7$				&	$99$\\
	\hline
	\hline
	\cite{CBIFV14}	& 	$83.5$				&	$93$					&	$93.3$				&	$81.7$				&	$93.8$				&	$97.2$\\
	%\hline
	\cite{NPB14}	& 	$81.5$				&	$72.1$				&	$64.3$				&	$44.2$				&	$68.9$				&	$90$\\
	%\hline
	\cite{SLWY14}	& 	$78.2$				&	$66.2$				&	$71.4$				&	$32.3$				&	$74.7$				&	$93.6$\\
	%\hline
	\cite{NF14}	& 	$81.5$				&	$73$					&	$67.1$				&	$45.2$				&	$68$					&	$88.9$\\
	%\hline
	\cite{FHWL14}	& 	$87$					&	$73$					&	$76$					&	$43$					&	$61$					&	$87$\\
	%\hline
	\cite{YWALH14}& 	$72$					&	$60$					&	$70$					&	$41$					&	$53$					&	$66$\\
	\hline
\end{tabular}
&
\begin{tabular}{|c|cccccc|}
	\hline
	Methods		&	CE					&	HO					&	CS					&	FS					&	NU					&	CY	\\
	\hline
	\hline
	FuzzySZM	& 	\textcolor{red}{$100$}	&	\textcolor{red}{$100$}	&	\textcolor{red}{$100$}	&	\textcolor{red}{$100$}	&	\textcolor{red}{$100$}	&	\textcolor{red}{$100$}\\
	SZM			& 	\textcolor{red}{$100$}	&	\textcolor{red}{$100$}	&	$98.6$				&	$86.6$				&	\textcolor{red}{$100$}	&	\textcolor{red}{$100$}\\
	\hline
	\hline
	FuzzyRLM	& 	\textcolor{red}{$100$}	&	\textcolor{red}{$100$}	&	\textcolor{red}{$100$}	&	\textcolor{red}{$100$}	&	\textcolor{red}{$100$}	&	\textcolor{red}{$100$}\\
	RLM			& 	\textcolor{red}{$100$}	&	$97.2$				&	\textcolor{red}{$100$}	&	$86.6$				&	$98.3$				&	\textcolor{red}{$100$}\\
	\hline
	\hline
	\cite{CBIFV14}	& 	$83$					&	\textcolor{red}{$100$}	&	\textcolor{red}{$100$}	&	\textcolor{red}{$100$}	&	\textcolor{red}{$100$}	&	\textcolor{red}{$100$}\\
	%\hline
	\cite{NPB14}	& 	$83.3$				&	\textcolor{red}{$100$}	&	$80$					&	\textcolor{red}{$100$}	&	$75$					&	\textcolor{red}{$100$}\\
	%\hline
	\cite{SLWY14}	& 	$83$					&	$80$					&	$80$					&	$20$					&	$75$					&	\textcolor{red}{$100$}\\
	%\hline
	\cite{NF14}	& 	\textcolor{red}{$100$}	&	\textcolor{red}{$100$}	&	$60$					&	$50$					&	\textcolor{red}{$100$}	&	\textcolor{red}{$100$}\\
	%\hline
	\cite{FHWL14}	& 	\textcolor{red}{$100$}	&	$60$					&	$80$					&	$50$					&	$50$					&	\textcolor{red}{$100$}\\
	%\hline
	\cite{YWALH14}& 	$83$					&	$60$					&	$80$					&	$75$					&	$75$					&	$75$\\
	\hline
\end{tabular}
\end{tabular}
}
\caption{Percentage predictions comparisons with the state-of-the-art at the cell level (left) and at the image level (right), using a neural network.}
\label{Table_ResultsICPR2012comp}
\end{table}

\begin{table}[!ht]
\resizebox{14cm}{!}{
\centering
\begin{tabular}{cc}
\begin{tabular}{|c|cccccc|}
	\hline
		&	CE								&	HO								&	CS								&	FS								&	NU								&	CY	\\
	\hline
	CE	&	\cellcolor{int93}{\textcolor{white}{$93.2$}}	&	\cellcolor{int01}{$1.1$}				&	\cellcolor{int04}{$3.9$}				&	\cellcolor{int01}{$1.1$}				&	$0$								&	$0.5$	\\
	HO	&	$0$								&	\cellcolor{int92}{\textcolor{white}{$92.1$}}	&	\cellcolor{int02}{$1.5$}				&	\cellcolor{int06}{$6$}					&	$0.3$							&	$0$		\\
	CS	&	\cellcolor{int01}{$0.9$}				&	$0$								&	\cellcolor{int93}{\textcolor{white}{$93.3$}}	&	\cellcolor{int02}{$1.9$}				&	\cellcolor{int01}{$1.4$}				&	\cellcolor{int01}{$2.3$}	\\
	FS	&	\cellcolor{int01}{$1.4$}				&	\cellcolor{int05}{$4.8$}				&	$0.4$							&	\cellcolor{int92}{\textcolor{white}{$92.3$}}	&	$0.4$							&	$0.4$	\\
	NU	&	$0.4$							&	\cellcolor{int01}{$0.8$}				&	$0$								&	\cellcolor{int01}{$0.8$}				&	\cellcolor{int98}{\textcolor{white}{$97.9$}}	&	$.0$		\\
	CY	&	$0$								&	$0$								&	$0$								&	$0$								&	$0$								&	\cellcolor{black}{\textcolor{white}{$100$}}	\\
	\hline
\end{tabular}
&
\begin{tabular}{|c|cccccc|}
	\hline
		&	CE								&	HO								&	CS								&	FS								&	NU								&	CY	\\
	\hline
	CE	&	\cellcolor{int93}{\textcolor{white}{$93.4$}}	&	$0.3$							&	\cellcolor{int02}{$1.7$}				&	\cellcolor{int01}{$1.4$}				&	\cellcolor{int02}{$2.2$}				&	$0$	\\
	HO	&	$0.3$							&	\cellcolor{int94}{\textcolor{white}{$94.2$}}	&	$0.3$							&	\cellcolor{int03}{$3.6$}				&	\cellcolor{int02}{$1.5$}				&	$0$		\\
	CS	&	\cellcolor{int02}{$1.9$}				&	$0$								&	\cellcolor{int94}{\textcolor{white}{$94.7$}}	&	\cellcolor{int01}{$1.4$}				&	$0.4$							&	\cellcolor{int01}{$1.4$}	\\
	FS	&	\cellcolor{int01}{$1.4$}				&	\cellcolor{int02}{$2.4$}				&	\cellcolor{int01}{$1.4$}				&	\cellcolor{int92}{\textcolor{white}{$92.7$}}	&	$0.5$							&	\cellcolor{int01}{$1.4$}	\\
	NU	&	$0.4$							&	\cellcolor{int01}{$1.2$}				&	\cellcolor{int01}{$0.8$}				&	$0$								&	\cellcolor{int97}{\textcolor{white}{$97.5$}}	&	$0$		\\
	CY	&	$0$								&	$0$								&	$0$								&	$0$								&	$0$								&	\cellcolor{black}{\textcolor{white}{$100$}}	\\
	\hline
\end{tabular}
%(a) & (b)
\end{tabular}
}
\caption{Confusion matrices for the results obtained at the cells level, using the FuzzySZM  (left) and the FuzzyRLM (right). The cell color is proportional to the value.}
\label{Table_ResultsICPR2012}
\end{table}

In the table~\ref{Table_ResultsICPR2012}, the fuzzy versions provide high prediction rates for each class. The methods produce efficient features describing each class without any ambiguity. This result is confirmed at the image level in the table~\ref{Table_ResultsICPR2012comp}, where we can observe that our classification is highly accurate. Moreover, the regular versions (RLM and SZM) provide comparable results as~\cite{CBIFV14}, but the fuzzy versions outperform for most of the classes at the cell level and the image level. From the same tables, we can confirm that the prediction rates for the cytoplasmatic and nucleolar classes are higher than other classes. This is due to these classes having typical textures different from the others: cytoplasmic cells are highly heterogeneous with a dark nucleus, and the nucleolar cells have big homogeneous bright patterns. Consequently, they appear atypical and easier to classify. For the same reasons, the fine speckled class has among the lowest predictions rates, because slightly speckled cells may appear homogeneous and more speckled cells may appear coarse speckled.

\subsection{ICIP 2013 contest and HPA datasets}
\label{SubSection_Results}

%The tables~\ref{Table_ResultsICIP},~\ref{Table_ResultsRF_HPA} and~\ref{Table_ResultsMLP_HPA} present results obtained on ICIP 2013 contest and HPA datasets. The blue numbers indicate when the fuzzy version improves the corresponding basic algorithm performances, and the red number points out the optimal performance for each class.
The table~\ref{Table_ResultsICIP} shows the results on the ICIP $2013$ contest dataset, which contains highly noisy images. We can observe that the fuzzy versions using the fuzzy zones significantly improve the performances for most classes. Indeed, the FuzzyRLM systematically surpasses the RLM, and the FuzzySZM surpasses the SZM for $75\%$ of cases, at both cell and image levels. Moreover, excepting only one case, the best result is provided by the fuzzy version.

\begin{table}[!ht]
\resizebox{14cm}{!}{
\begin{tabular}{cc}
\begin{tabular}{|l|cccccc|}
	\hline
				&	CE					&	GO					&	HO					&	NU					&	NM					&	SP\\
	\hline
	\hline
	FuzzySZM	&	\textcolor{red}{$97.04$}	&	$86.99$				&	\textcolor{red}{$95.75$}	&	\textcolor{blue}{$93.25$}	&	$91.86$				&	$92.27$\\
	SZM			&	$95.71$				&	\textcolor{red}{$89.92$}	&	$94.79$				&	$92.49$				&	$92.26$				&	$92.5$\\
	%FSZM		&	$88.34$				&	$70.3$				&	$89.65$				&	$87.8$				&	$86.05$				&	$80.17$\\
	\hline
	\hline
	FuzzyRLM	&	\textcolor{blue}{$95.71$}	&	\textcolor{blue}{$88.89$}	&	\textcolor{blue}{$93.16$}	&	\textcolor{red}{$93.51$}	&	\textcolor{red}{$92.59$}	&	\textcolor{red}{$93.05$}\\
	RLM			&	$92.7$				&	$72.46$				&	$90.5$				&	$87.43$				&	$86.16$				&	$86.78$\\
	%FRLM		&	$91.29$				&	$72.45$				&	$88.94$				&	$85.03$				&	\textcolor{blue}{$86.59$}	&	$84.67$\\
	\hline
	\hline
	COM			&	$95.7$				&	$86.27$				&	$93.53$				&	$91.88$				&	$91.1$				&	$91.71$\\
	%FCOM		&	$94.19$				&	$79.47$				&	$92.04$				&	$90.62$				&	$88.09$				&	$89.42$\\
	\hline
\end{tabular}
&
\begin{tabular}{|l|cccccc|}
	\hline
				&	CE					&	GO					&	HO					&	NU					&	NM					&	SP\\
	\hline
	\hline
	FuzzySZM	&	\textcolor{red}{$96.57$}	&	\textcolor{blue}{$85.96$}	&	\textcolor{red}{$95.75$}	&	\textcolor{red}{$92.96$}	&	\textcolor{blue}{$91.73$}	&	\textcolor{red}{$91.24$}\\
	SZM			&	$92.84$				&	$85.86$				&	$90.13$				&	$87.33$				&	$89.35$				&	$87.53$\\
	%FSZM		&	$88.34$				&	$67.99$				&	$86.22$				&	$81.04$				&	$84.1$				&	$80.17$\\
	\hline
	\hline
	FuzzyRLM	&	\textcolor{blue}{$93.95$}	&	\textcolor{red}{$90.16$}	&	\textcolor{blue}{$93.51$}	&	\textcolor{blue}{$89.49$}	&	\textcolor{red}{$92.87$}	&	\textcolor{blue}{$89.15$}\\
	RLM			&	$91.86$				&	$79.51$				&	$90.33$				&	$87.83$				&	$87.88$				&	$85.28$\\
	%FRLM		&	$90.69$				&	$77.86$				&	$89.42$				&	$83.03$				&	$85.81$				&	$82.91$\\
	\hline
	\hline
	COM			&	$95.76$				&	$90.68$				&	$92.26$				&	$88.8$				&	$88.38$				&	$89.42$\\
	%FCOM		&	$94.48$				&	$82.27$				&	$91.69$				&	\textcolor{blue}{$88.94$}	&	$86.99$				&	$87.39$\\
	\hline
\end{tabular}
\end{tabular}
}
\caption{Percentage predictions comparisons over the different statistical matrices on the ICIP 2013 contest dataset, obtained with random forests (left) and neural network (right).}
\label{Table_ResultsICIP}
\end{table}

\begin{table}[!ht]
\resizebox{\columnwidth}{!}{
\begin{tabular}{|l|ccccccccccc|}
	\hline
				&	CE					&	CY					&	CK					&	ER					&	GO					&	MI					&	NU					&	NI						&	NIwo					&	PL					&	VE\\
	\hline
	\hline
	FuzzySZM	&	\textcolor{blue}{$62.12$}	&	\textcolor{red}{$86.5$}	&	\textcolor{red}{$71.57$}	&	\textcolor{red}{$67.2$}	&	$62.55$				&	\textcolor{red}{$77.97$}	&	\textcolor{blue}{$59.25$}	&	\textcolor{red}{$73.57$}			&	\textcolor{red}{$82.3$}	&	\textcolor{blue}{$55.64$}	&	$69.24$\\
	SZM			&	$60.49$				&	$80.64$				&	$64.59$				&	$63.42$				&	$64.73$				&	$77.69$				&	$58.83$				&	$73.35$					&	$80.69$				&	$54.13$				&	\textcolor{red}{$76.66$}\\
	\hline
	\hline
	FuzzyRLM	&	$61.95$				&	\textcolor{blue}{$82.05$}	&	$64.49$				&	\textcolor{blue}{$66.75$}	&	$60.69$				&	$73.48$				&	\textcolor{red}{$62.05$}	&	$71.18$					&	$80.21$				&	\textcolor{red}{$56.9$}	&	$69.27$\\
	RLM			&	\textcolor{red}{$65.14$}	&	$80.99$				&	$65.67$				&	$64.2$				&	\textcolor{red}{$66.23$}	&	$74.66$				&	$59.44$				&	$72.67$					&	$80.82$				&	$56.7$				&	$71.27$\\
	\hline
	\hline
	COM			&	$63.51$				&	$81.83$				&	$62.96$				&	$61.04$				&	$62.87$				&	$74.55$				&	$61.36$				&	$73.05$					&	$79.48$				&	$55.41$				&	$73.65$\\
	\hline
\end{tabular}
}
\caption{Percentage prediction results using random forest on HPA dataset.}
\label{Table_ResultsRF_HPA}
\end{table}

\vspace{0.5cm}

\begin{table}[!ht]
\resizebox{\columnwidth}{!}{
\begin{tabular}{|l|ccccccccccc|}
	\hline
				&	CE					&	CY					&	CK					&	ER					&	GO					&	MI					&	NU					&	NI						&	NIwo					&	PL					&	VE\\
	\hline
	\hline
	FuzzySZM	&	\textcolor{red}{$87.55$}	&	\textcolor{red}{$88.32$}	&	$83.27$				&	\textcolor{red}{$85.74$}	&	\textcolor{red}{$81.7$}	&	\textcolor{red}{$84.88$}	&	\textcolor{blue}{$67.2$}	&	$77.89$					&	\textcolor{red}{$82.58$}	&	$64.29$				&	$74.64$\\
	SZM			&	$78.52$				&	$86.86$				&	$83.72$				&	$84.34$				&	$74.49$				&	$82.38$				&	$60.08$				&	\textcolor{red}{$79.08$}			&	$80.27$				&	$68$					&	\textcolor{red}{$82.11$}\\
	\hline
	\hline
	FuzzyRLM	&	$78.64$				&	$87.55$				&	\textcolor{red}{$84.51$}	&	\textcolor{blue}{$85.11$}	&	\textcolor{blue}{$79.87$}	&	$76.8$				&	$62.67$				&	$69.56$					&	$79.93$				&	\textcolor{red}{$69.41$}	&	$73.39$\\
	RLM			&	$81.63$				&	$87.34$				&	$83.06$				&	$84.86$				&	$79.38$				&	$81.93$				&	\textcolor{red}{$69.91$}	&	$78.3$					&	$80.37$				&	$65.63$				&	$81.56$\\
	\hline
	\hline
	COM			&	$76.66$				&	$88.01$				&	$80.86$				&	$82.19$				&	$80.54$				&	$78.87$				&	$66.95$				&	$74.11$					&	$79.44$				&	$62.98$				&	$79.01$\\
	\hline
\end{tabular}
}
\caption{Percentage prediction results using neural network classification results on HPA dataset.}
\label{Table_ResultsMLP_HPA}
\end{table}

The tables \ref{Table_ResultsRF_HPA} and \ref{Table_ResultsMLP_HPA} present results obtained on the HPA dataset, which contains high quality (staining, illumination, contrast, etc.) images. The results are less dramatic, because the FuzzyRLM does not improve performances in most cases. However the FuzzySZM still performs as well as SZM if not better.

\section{Conclusion and Perspectives}
\label{Section_Conclusion}

This paper presents different versions of fuzzy statistical matrices. The first version is a generalization of an existing technique, and introduces the fuzzification at the matrix filling level by spreading the information. The results presented in section \ref{Section_Results} show that this method never improved the results for the three datasets used in this paper. Even if this method was introduced to reduce noise sensitivity, the results are lower than the classical algorithm.\\
Next we define the original fuzzy zone, which is not flat but has fuzzy values. The fuzzy zones are used to fill statistical matrices, and then to create fuzzy statistical matrices. These new matrices are powerful descriptors, particularly effective at characterizing highly noisy images. The efficiency is particularly significant for the fuzzy run length matrix, which systematically outperforms the regular run length matrix performances, on both noisy datasets and using different classification methods. Moreover, the fuzzy size zone matrix using fuzzy zones also provides good characteristics on high quality images. In order to validate the results, we performed a comparison with the best methods from the state-of-the-art, which provide comparable results with the regular matrices, but are outperformed by the new fuzzy versions.\\
As a result this paper demonstrates that the new fuzzy version using fuzzy zones generates reliable and effective fuzzy statistical matrices, and provides better results than the original fuzzy version. Moreover, the new fuzzy statistical matrices systematically provide better results than the widely used co-occurrences matrix. Therefore our methods can be used to improve the characterization of images, for example medical imaging and the delicate issue of describing cancerous cells~\cite{KSKB08} or tumors~\cite{DKB06,HWFJ12}.\\
The classic statistical matrices and the new fuzzy statistical matrices use different gray level reduction algorithms and quantizations. Unfortunately, no fine-tuning method exists to automatically determine the optimal configuration. Moreover, the experiments perform in this paper have shown that the performances greatly vary according to the dataset: no gray level reduction algorithm or quantization has proven to be more likely to provide better results. Consequently, it is necessary to test a maximum of configurations in order to find the best results.

\section*{Acknowledgments}
This work was funded by NSF award 1027834. Any opinions, findings, conclusions or recommendations expressed in this publication are those of the authors and do not reflect the views of the NSF.

\end{document}